\documentclass{article}

\usepackage[utf8]{inputenc} 
\usepackage[T1]{fontenc}    
\usepackage{hyperref}       
\usepackage{url}            
\usepackage{booktabs}       
\usepackage{amsfonts}       
\usepackage{nicefrac}       
\usepackage{microtype}      
\usepackage{xcolor}         
\usepackage{graphicx}
\usepackage{verbatim} 
\usepackage{amsmath}
\usepackage{float}
\usepackage{geometry}
\title{Confronting Discrimination in Classification: Smote Based on Marginalized Minorities in the Kernel Space for Imbalanced Data}

\author{%
  Lingyun Zhong \\
  Department of Civil and Environmental Engineering\\
  University of Michigan\\
  Ann Arbor, MI 48109 \\
  \texttt{lingyunz@umich.edu} \\
}

\begin{document}

\maketitle

\begin{abstract}
    Financial fraud detection poses a typical challenge characterized by class imbalance, where instances of fraud are extremely rare but can lead to unpredictable economic losses if misidentified. Precisely classifying these critical minority samples represents a challenging task within the classification. The primary difficulty arises from mainstream classifiers, which often exhibit "implicit discrimination" against minority samples in evaluation metrics, which results in frequent misclassifications, and the key to the problem lies in the overlap of feature spaces between majority and minority samples. To address these challenges, oversampling is a feasible solution, yet current classical oversampling methods often lack the necessary caution in sample selection, exacerbating feature space overlap. In response, we propose a novel classification oversampling approach based on the decision boundary and sample proximity relationships. This method carefully considers the distance between critical samples and the decision hyperplane, as well as the density of surrounding samples, resulting in an adaptive oversampling strategy in the kernel space. Finally, we test the proposed method on a classic financial fraud dataset, and the results show that our proposed method provides an effective and robust solution that can improve the classification accuracy of minorities.
\end{abstract}

\section{Introduction}
The class imbalance problem is a classic classification problem, which arises because the number of negative samples (i.e., majority class) in the data set is much larger than the number of positive samples (i.e., minority class)\cite{lemaitre2017imbalanced}. This type of problem is common in many fields. For example, in the field of financial fraud, the occurrence of occasional small-probability fraud will cause huge economic losses. Therefore, accurately identifying positive samples will be the key to the class imbalance problem.

The first difficulty in the class imbalance problem is mainly due to the rarity of positive samples, which has two connotations\cite{sasada2020resampling}: One is \textbf{absolutely rare}, which makes the data not representative enough and has a lot of noise; the other is \textbf{relatively rare}, which causes the feature space to overlap seriously, making it hard for the model to accurately separate the two classes. The second reason is the potential discrimination toward positive samples by current mainstream classifiers. Many current models treat the majority and minority classes equally when evaluating classification accuracy, resulting in the direction of model evaluation being naturally biased towards the majorities; the third reason is the potential discrimination toward important samples in positive samples by the oversampling model. SMOTE, as a classic oversampling method to solve class imbalance\cite{10.5555/1622407.1622416}, only selects the data randomly when expanding the minorities, which may result in more serious feature space overlap because of the ignoration of important samples in minorities.

To solve the various problems mentioned above, we propose a hierarchical Smote Based on Marginalized Minorities(MM-SMOTE). First, we use the basic SVM classifier to roughly classify the data, and obtain the support vectors in minorities as important samples for sampling; then assign weights to those support vectors based on their distance to the decision hyperplane; and then based on the k-nearest neighbors of support vectors, we used an adaptive oversampling to generate synthetic samples; finally, synthetic samples are used to augment the original kernel function of the basic SVM to form a new classifier.

In summary, our work makes the following contributions:
\begin{itemize}
    \item {Proposing a new sampling sample selection method, which can provide a theoretical basis for selecting minority samples in oversampling methods.}
    \item {Designing an adaptive oversampling algorithm to generate synthetic samples, which is based on neighbor density analysis of minority samples and effectively reduces the bias of the decision hyperplane obtained on imbalanced data sets.}
    \item {Introducing k-means combined with a random undersampling method to form financial fraud data sets with different imbalance ratios. It can be proved that the method we proposed has good stability under different imbalance ratios.}
\end{itemize}
The rest of the paper is organized as follows. Section \ref{related work} makes an overview of several methodologies to handle imbalance classification, respectively.
Section \ref{methodology} presents the proposed method in detail. Section \ref{experiment} shows
the experimental results of the proposed method in comparison
with the existing counterparts. Finally, we draw a conclusion in
Section \ref{conclusion}.
\section{Related Work}\label{related work}

\subsection{Imbalanced Classification at Data Preprocessing Level}
The basic idea of the imbalanced data processing method at the data preprocessing level is to change the sample distribution of the data set, thereby reducing or eliminating the imbalance.

Undersampling reduces the imbalance of the data set by deleting old majority-class samples. Random undersampling\cite{mishra2017handling} is a representative type of undersampling algorithm. It achieves the purpose of balancing the majority samples and the minority samples by randomly selecting points in the majority samples and deleting them from the data set. However, there is a problem with random undersampling, which is that it ignores the changes in the distribution of data that may be caused by randomly deleting data points. Therefore, some researchers\cite {lin2017clustering} consider combining random undersampling with clustering algorithms such as k-means to form several majority class samples clusters. Representative data points are sampled in each cluster using random undersampling, which ensures the stability of the data distribution.

Oversampling reduces the imbalance of the data set by adding new minority-class samples. The SMOTE algorithm\cite{10.5555/1622407.1622416} is one of the most representative oversampling methods. This algorithm generates new minority class samples without duplication between minority class samples through linear interpolation. Since the SMOTE algorithm is too random, cannot control the position of synthetic samples, and is susceptible to noise interference, researchers have further proposed a series of improved oversampling methods including Borderline-SMOTE\cite{han2005borderline}, Kmeans-SMOTE\cite{douzas2018improving}, etc. The difference between Borderline-SMOTE and SMOTE is that it first finds minority class samples near the decision boundary based on rule judgment, and then selectively generates new samples. Kmeans-SMOTE uses K-Means to cluster minority class samples and perform SMOTE in the clusters to avoid noise while eliminating inter-class imbalance and intra-class imbalance. In addition to oversampling the original data, a method called WK-SMOTE\cite{mathew2017classification} is conducted, which is based on the kernel function and implicitly oversamples the data in the kernel space, which can effectively improve the classification accuracy.

\subsection{Imbalanced Classification at Algorithm Level}
The most mainstream algorithm improvement for class imbalance focuses on the improvement of the loss function or objective function. It uses a simple idea to set a relatively larger weight for the misclassification of minority samples\cite{fumera2002cost}, which means that the classifier error classifying minority samples imposes more penalties, making the basic classifier more biased toward minority samples, thereby achieving class balance in an algorithmic sense. The advantage of this method is that it simply introduces a hyperparameter of a penalty coefficient and does not increase the complexity of the solution. Based on this idea, researchers are not satisfied with giving the same weight to all minority samples. Because of possible noise samples in the minority samples, it is unfair to assign the same penalty coefficient to all samples\cite{huang2005weighted}. Therefore, the researchers try to give each sample point a different penalty coefficient for misclassification errors, achieving cost-sensitive learning at a more micro level.

However, the disadvantage of cost-sensitive learning is that when the data set is extremely unbalanced, even if the weight or penalty term is set, the information learned is still very limited. In this direction, sampling methods are still needed

\section{Methodology}\label{methodology}
As mentioned in the literature review, randomly duplicating samples from the initial dataset during oversampling can exacerbate class overlapping. Addressing this concern, we propose a hierarchical classification approach. Initially, we establish a fundamental decision hyperplane through an initial SVM classifier and identify crucial samples within the minorities (i.e., support vectors). Subsequently, we weigh these important samples based on their distances to the decision hyperplane. Following this, we further group these important samples based on their proximity relationships, adaptively oversampling to create a synthetic dataset. Finally, by employing kernel tricks we augment the original kernel matrix and reintroduce it into the SVM to obtain a new classifier. Details of the method will be elaborated on in the following sections.
\subsection{Key Sample Identification}
There is no doubt that in the classification problem based on support vector machines, what deserves our attention, and the point that has an important impact on classification must be those support vectors.

The basic optimization equation of the soft margin SVM model is as follows according to formulation \ref{con:Soft SVM}:
\begin{equation}
\begin{aligned}
& \underset{w, b, \xi_i}{\text{minimize}}\label{con:Soft SVM}
& & \frac{1}{2} \|w\|^2 + C \sum_{i=1}^{N} \xi_i \\
& \text{subject to}
& & y_i (w^T x_i + b) \geq 1 - \xi_i, \quad i = 1, 2, ..., N \\
&&& \xi_i \geq 0, \quad i = 1, 2, ..., N,
\end{aligned}
\end{equation}
Where \( w \in \mathbb{R}^d \) be a normal vector; \( b \in \mathbb{R} \) be a bias; \( \xi_i \) be a slack variable; \( C \) represents the regularization parameter; \( x_i \) denotes the data sample; the class label \( y_i \) of \( x_i \) is such that \( y_i \in \{1, -1\} \).

We can transform it into its corresponding dual optimization problem. So the support vector machine is the classifier obtained by solving the optimization problem \ref{con:dual}:
\begin{equation}
\begin{aligned}\label{con:dual}
& \underset{\boldsymbol{\alpha}}{\text{maximize}}
& & \sum_{i=1}^{N} \alpha_i - \frac{1}{2} \sum_{i=1}^{N} \sum_{j=1}^{N} \alpha_i \alpha_j y_i y_j \mathbf{x}_i^T \mathbf{x}_j \\
& \text{subject to}
& & \sum_{i=1}^{N} \alpha_i y_i = 0 \\
& & & 0 \leq \alpha_i \leq C, \quad i = 1, \ldots, N.
\end{aligned}
\end{equation}
where $\alpha_i$ represents the Lagrange multiplier with respect to $x_i$, Solving it is equivalent to finding a solution $\alpha$ satisfying the Karush-Kuhn-Tucker (KKT) conditions if the regularity condition holds.

According to the classification results of the support vector machine, we can get optimal slack variables satisfy:
\begin{equation}
\begin{aligned}\label{con:slack}
\xi_i^{*}=max(0,1-y_i((\omega^*)^Tx+b^*))
\end{aligned}
\end{equation}
where $\alpha^*$ denote a dual optimal solution;  $\omega^*$=$\sum\alpha_i^*y_ix_i$ and $b^*$=$y_j-\sum \alpha^*y_i\kappa(x_i,x_j)$.

\begin{figure}[H]
	\centering
	\includegraphics[width=0.5\linewidth]{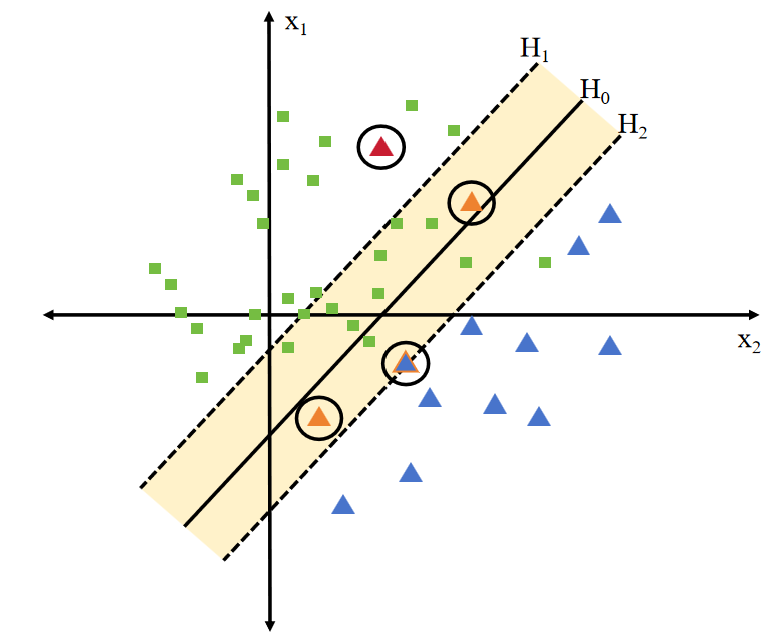}\label{fig:support v}
	\caption{SVM classification samples. $H_0$ represents the decision hyperplane;$H_1$ and $H_2$ represent hyperplanes generalized by majorities and minorities. The points represented by blue triangles are safe points that are correctly classified among minorities. Points located in the yellow margin area and its boundaries are points that may be misclassified; red points represent points that will definitely be misclassified. , the latter two points are support vectors, which are also the focus of this study.}
	\label{chutian}
\end{figure}

Therefore, after using the basic classifier SVM to roughly classify the original data set, we can see the following four cases in minorities according to Figure \ref{fig:support v}:
\begin{itemize}
    \item {if $y_i(\langle \omega^*, x_i \rangle+b^*) > 1$, then $\xi_i^*=0$, $x_i$ is not a support vector and it is safe}
    \item {if $y_i(\langle \omega^*, x_i \rangle+b^*) = 1$, then $\xi_i^*=0$, $x_i$ is a support vector}
    \item {if $y_i(\langle \omega^*, x_i \rangle+b^*) < 1$, then $\xi_i^*>0$, $x_i$ is a support vector}
    \item {if $y_i(\langle \omega^*, x_i \rangle+b^*) < 0$, then $\xi_i^*>1$, $x_i$ is a support vector}
\end{itemize}

We don't need to be overly concerned about safe points in minorities since they will consistently be correctly classified. If oversampling is exclusively based on those safe points, the resulting points will be significantly distant from the decision hyperplane. This situation is not beneficial for achieving a clearer classification between majorities and minorities. It is more pertinent to focus on points located on the decision hyperplane, those prone to misclassification or currently misclassified. Oversampling these points is likely to refine the original decision boundaries and effectively make the classification.

\subsection{Key Sample Selection}
Although we have identified those important samples that deserve our attention, it is obvious that these important samples are not completely equally important. Those points that are closer to the decision hyperplane often mean that Points that are “more difficult to learn”\cite{barua2012mwmote}, that is, these points are key parts of the marginal groups of minorities. By paying more attention to them, we may be able to more clearly find features that are difficult to separate.

To achieve the above goal, we need to assign a weight to each marginal-important sample\cite{ren2023grouping}. This weight will be used as the probability of being selected. The greater the weight of the important sample, the greater the probability of being selected. The calculation formula of the weight \ref{con:weight} is as follows:
\begin{equation}
W_{sv^+}(x_i) = \frac{e^{-|L(x_i)|}}{\sum_{x_k \in sv^+} e^{-|L(x_k)|}}\label{con:weight}
\end{equation}
where $L(x_i)$ represents the distance between minority important samples $sv^+$ and the decision hyperplane, the closer it is to the decision hyperplane, the greater the probability of being selected (that is, the greater the weight).

\subsection{Synthetic Sample Generation}
Through the above method, we assigned a weight to each important sample. However, a certain limitation still exists. Samples that are relatively close to the decision hyperplane pose a significant challenge to separate from the majorities during the original learning process. Increasing the weight of these samples might lead to the accumulation of synthetic samples around the decision hyperplane, thereby expanding the class overlapping. To address this issue, drawing inspiration from the concept of Borderline-SMOTE\cite{han2005borderline}, we take into account the proximity relationship of important samples and perform an adaptive oversampling on them.

\begin{figure}[H]
	\centering
	\includegraphics[width=0.9\linewidth]{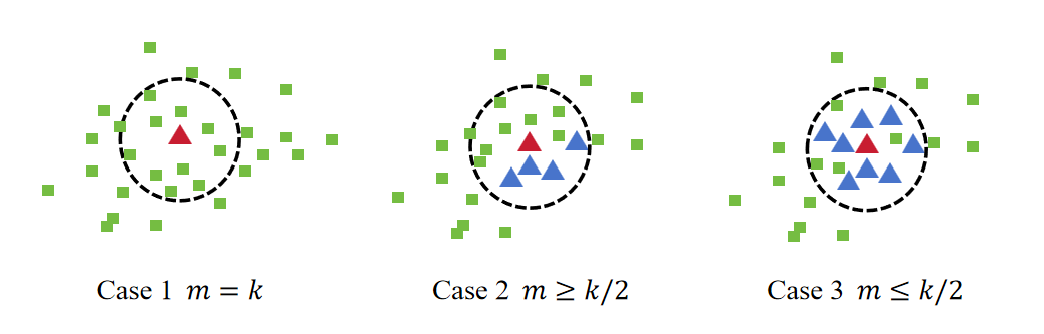}\label{fig:3 case}
	\caption{3 cases of support vectors and their $k$ nearest neighbors. The red triangles represent the selected sampling points, the blue triangles represent the surrounding minority class samples, and the green squares represent the surrounding majority class samples.}
	\label{chutian}
\end{figure}

We initially conduct $k$-nearest neighbor sampling on the selected important samples from the previous step to acquire label information for their surrounding sample points. Three potential scenarios(labels) are illustrated in Figure \ref{fig:3 case}. We document the count of surrounding majority class sample points as $m$.

For case 1, if $m$=$k$, this sample point is entirely encircled by majority class samples; hence, it will be treated as noise in the minority class samples.

For case 2, if $m \geq k/2$, it can be observed that the selected sample is predominantly surrounded by majority class samples. Therefore, overall, this sample is relatively risky. When oversampling it, we will allow a more conservative sampling approach to keep it as close as possible to the decision hyperplane. The formula is shown in \ref{con:k.21}:
\begin{equation}
\begin{aligned}\label{con:k.21}
\phi(x_{ij})=\phi(x_i) + \delta \left( \phi(x_j) - \phi(x_i) \right), \delta \in (0,1)
\end{aligned}
\end{equation}
For case 3, if $m \leq k/2$, it can be seen that the selected sample is predominantly surrounded by the minority class samples. Therefore, overall, this sample is relatively safe. When oversampling it, we will allow a more aggressive sampling approach to keep it away from the decision hyperplane to a certain extent. The formula is shown in \ref{con:k.22}:
\begin{equation}
\begin{aligned}\label{con:k.22}
\phi(x_{ij})=\phi(x_i) + \delta \left( \phi(x_j) - \phi(x_i) \right), \delta \in (-1,0)
\end{aligned}
\end{equation}
where \(\phi(\cdot)\) is a nonlinear transformation to map all of the samples from the input space to a linearly separable feature space.

According to our proposed method, by adaptively oversampling based on the density distribution of surrounding samples, it is effective in expanding minority class samples to regions where the density of majorities is relatively low, thereby adjusting the decision hyperplane.

\subsection{Kernel Matrix Augment}

We set the size of the original data set as $n$, and the size of the synthetic data set as $s$. According to the fundamental classifier SVM, we can get that kernel matrix $K_1$=$(K^{1}_{p,q}) \in \mathbb{R}^{n\times n}$ obtained from the original data set, where the element of kernel matrix is $K^{1}_{p,q}=\kappa(x_p,x_q)$, and $\kappa(\cdot,\cdot)$ denotes a kernel function equivalent to the inner product between two samples in a specified feature space.

After generating $s$ synthetic minorities in feature space, we can augment $K^1$ via the following equation \ref{con:Matrix0}:
\begin{equation}
\begin{aligned}\label{con:Matrix0}
K&=
\left [ \begin{array} {c | c c}
K^1 & K_{+}^2 & K_{-}^2 \\ \hline
\\
(K_{+}^2)^T & K_{+,+}^3 & K_{+,-}^3  \\ 
\\
(K_{-}^2)^T & K_{-,+}^3 & K_{-,-}^3 
\end{array}\right ]\\
\\
&=\begin{bmatrix} K^1 & K^2 \\ (K^2)^T & K^3 \end{bmatrix}
\end{aligned}
\end{equation}

where $K^2=\begin{bmatrix} K_{+}^2 & K_{-}^2\end{bmatrix} \in \mathbb{R}^{n \times s}$ with element $K^2_{p,q}=\kappa(x_p,\hat{x}_q)$, $x_p \in D^{ori}$(i.e., the original data set) and $\hat{x}_q \in D^{syn}$(i.e., the synthetic smaple set). $K^3=\begin{bmatrix} K_{+,+}^3 & K_{+,-}^3  \\K_{-,+}^3 & K_{-,-}^3 \end{bmatrix}\in \mathbb{R}^{s \times s}$ with element $K^3_{p,q}=\kappa(\hat{x}_p, \hat{x}_q)$, where $\hat{x}_p$, $\hat{x}_q$ $\in D^{syn}$. When the kernel function subscript is $+$, it means that $\hat{x}_q$ belongs to Case 2, when the kernel function subscript is $-$, it means that $\hat{x}_q$ belongs to Case 3. 

Each element in $K^2$ is the inner product of the original data set sample $x_p$ and the synthetic sample $\hat{x}_{ij}$ in the feature space using the following equation \ref{con:K1}:
\begin{gather}
\begin{split}\label{con:K1}
    \kappa(x_p, \hat{x}_{ij})&=\langle \phi(x_p), \phi(x_{ij}) \rangle\\
    &=\phi(x_p)^T [\phi(x_i) + \delta \left( \phi(x_j) - \phi(x_i) \right)]\\
    &=(1 - \delta) \kappa(x_p, x_i) + \delta \kappa(x_p, x_j)
\end{split}
\end{gather}
where $\delta \in (0,1)$ when $x_i$ belongs to case 2; $\delta \in (-1,0)$ when $x_i$ belongs to case 3.

The elements in $K^3$ are the inner product of two synthetic samples in the feature space and are calculated by equation \ref{con:k2}:
\begin{gather}
\begin{split}\label{con:k2}
    \kappa(\hat{x}_{lr}, \hat{x}_{ij})&=\langle \phi_{lr}, \phi_{ij} \rangle\\
    &=[\phi(x_l) + \delta_1 (\phi(x_r) - \phi(x_l))]^T[\phi(x_i) + \delta_2 (\phi(x_j) - \phi(x_i))]\\
    &=(1 - \delta_2)(1 - \delta_1) \kappa(x_i, x_l) + (1 - \delta_2) \delta_1 \kappa(x_i, x_r)\\
    &+ \delta_2(1 - \delta_1) \kappa(x_j, x_l) + \delta_2 \delta_1 \kappa(x_j, x_r)
\end{split}
\end{gather}

where $\delta_1,\delta_2 \in (0,1)$ when $x_i$, $x_l$ belongs to case 2; $\delta_1,\delta_2 \in (-1,0)$ when $x_i$, $x_l$ belongs to case 3.

Then, we can formulate a new decision function shown in equation \ref{con:decision function}  based on SVM using the augmented kernel matrix:
\begin{equation}
\begin{aligned}\label{con:decision function}
f(x)=sign(\sum_{i=1}^{D^{ori}}\alpha^{*}_i K(\textbf{x},x_i)+\sum_{i=1}^{D^{syn^+}}\alpha^{*}_i K(\textbf{x},x^{+}_{p})+\sum_{i=1}^{D^{syn^-}}\alpha^{*}_i K(\textbf{x},x^{-}_{p})+b^{*})
\end{aligned}
\end{equation}
where $\textbf{x}$ is the test data set; $x_i \in D^{ori}$; $x^{+}_p \in D^{syn^+}$(Case 2); $x^{-}_p \in D^{syn^{-}}$(Case 3); $\alpha^{*}_i$ and $b^{*}$ are mentioned before.

\section{Experiments}\label{experiment}
In this section, we first introduce datasets we use to evaluate MM-SMOTE and baseline models. Then we provide a detailed analysis to show the effectiveness of MM-SMOTE and its strong applicability to real-world financial fraud datasets.

\subsection{Dataset}
In this paper, we adopt the credit card fraud dataset from the Kaggle Competition: Credit Card Fraud Detection. The dataset contains transactions made by credit cards in September 2013 by European cardholders. This dataset presents transactions that occurred in two days, where we have 492 frauds out of 284,807 transactions. The dataset is highly unbalanced, the positive class (frauds) accounts for 0.172$\%$ of all transactions. It contains only numerical input variables which are the result of a PCA transformation. Feature 'Class' is the response variable and it takes value 1 in case of fraud and 0 otherwise. 
\subsection{Baseline Methods}
The proposed MM-SMOTE was evaluated against the following four baseline algorithms: (1)Traditional SVM classifier: used on data set without additional oversampling or weighting; (2) Class-weighted SVM classifier: Penalty weights are set for all samples based on the basic SVM model; (3) RUS-SVM: This is an approach that aims to balance the class distribution by randomly eliminating majorities. The modified data set is then used to train an SVM classifier; (4) SMOTE-SVM: Based on the random oversampling algorithm, new samples are artificially synthesized based on the minorities. The modified data set is then used to train an SVM classifier.
\subsection{Experimental Settings}
To comprehensively evaluate the effect of imbalance learning and eliminate the discrimination of evaluation metrics against minorities, we choose to use four metrics: \textbf{Precision}, \textbf{Recall}, \textbf{F1-score}, and \textbf{G-mean} to evaluate our algorithm and baselines. At the same time, to evaluate the stability of our algorithm under different imbalance ratios, we used k-means combined with random under sampling to extract subsets from majorities to form data sets with different balance ratios including 2:1, 4:1, 6:1, 8:1, 10:1, 70:1.
\subsection{Result Anaysis}
According to Table \ref{tab:ratio2}, \ref{tab:ratio4}, \ref{tab:ratio6}, \ref{tab:ratio8}, \ref{tab:ratio10}, \ref{tab:ratio70}, 
through comparisons at the same imbalance ratio, it becomes evident that our proposed method could balance prediction accuracy and recall rate. The resultant F1-score and G-mean markedly are greater than other baseline methods, underscoring the exceptional performance of our approach in addressing imbalance classification. Furthermore, when examining various imbalance ratios, our method exhibits remarkable stability as the imbalance ratio increases. The algorithm consistently maintains high accuracy in distinguishing between positive and negative samples.
\begin{table}[H]
    \centering
    \caption{Performance comparison on imbalanced datasets with 2:1 ratio}
    \begin{tabular}{|c|c|c|c|c|}
        \hline
        Algorithm & Precision & Recall & F1-score & G-mean \\
        \hline
        SVM & 0.9998 & 0.7198 & 0.8370 & 0.8483 \\
        \hline
        Class-weighted SVM & 0.8956 & 0.9082 & 0.9019 & 0.9019\\
        \hline
        RUS-SVM & 0.8986 & 0.9034 & 0.9010 & 0.9010 \\
        \hline
        SMOTE-SVM & 0.9123 & 0.9034 & 0.9078 &  0.9078 \\
        \hline
        MM-Smote & 0.8983 & 0.9103 & 0.9056 & 0.9057 \\
        \hline
    \end{tabular}
    \label{tab:ratio2}
\end{table}
\begin{table}[H]
    \centering
    \caption{Performance comparison on imbalanced datasets with 4:1 ratio }
    \begin{tabular}{|c|c|c|c|c|}
        \hline
        Algorithm & Precision & Recall & F1-score & G-mean \\
        \hline
        SVM & 0.9833 & 0.8115 & 0.8893 & 0.8933 \\
        \hline
        Class-weighted SVM & 0.9696 & 0.8744 & 0.9150 & 0.9160 \\
        \hline
        RUS-SVM & 0.8930 & 0.9130 & 0.9029 & 0.9030 \\
        \hline
        SMOTE-SVM & 0.9477 & 0.8841 & 0.9148 & 0.9153 \\
        \hline
        MM-Smote & 0.9394 & 0.8986 & 0.9185 & 0.9187 \\
        \hline
    \end{tabular}
    \label{tab:ratio4}
\end{table}
\begin{table}[H]
    \centering
    \caption{Performance comparison on imbalanced datasets with 6:1 ratio }
    \begin{tabular}{|c|c|c|c|c|}
        \hline
        Algorithm & Precision & Recall & F1-score & G-mean \\
        \hline
        SVM & 0.9950 & 0.7536 & 0.8577 & 0.8659 \\
        \hline
        Class-weighted SVM & 0.9691 & 0.8406 & 0.9003 & 0.9026 \\
        \hline
        RUS-SVM & 0.8992 & 0.8986 & 0.8989 & 0.8989 \\
        \hline
        SMOTE-SVM & 0.9575 & 0.8889 & 0.9219 & 0.9226 \\
        \hline
        MM-Smote & 0.9617 & 0.8937 & 0.9264 & 0.9270 \\
        \hline
    \end{tabular}
    \label{tab:ratio6}
\end{table}
\begin{table}[H]
    \centering
    \caption{Performance comparison on imbalanced datasets with 8:1 ratio}
    \begin{tabular}{|c|c|c|c|c|}
        \hline
        Algorithm & Precision & Recall & F1-score & G-mean \\
        \hline
        SVM & 0.9994 & 0.7343 & 0.8466 & 0.8567 \\
        \hline
        Class-weighted SVM & 0.9821 & 0.7923 & 0.8770 & 0.8821 \\
        \hline
        RUS-SVM & 0.9374 & 0.9082 & 0.9226 & 0.9227 \\
        \hline
        SMOTE-SVM & 0.9666 & 0.8647 & 0.9128 & 0.9143 \\
        \hline
        MM-Smote & 0.9547 & 0.8889 & 0.9206 & 0.9212 \\
        \hline
    \end{tabular}
    \label{tab:ratio8}
\end{table}
\begin{table}[H]
    \centering
    \caption{Performance comparison on imbalanced datasets with 10:1 ratio }
    \begin{tabular}{|c|c|c|c|c|}
        \hline
        Algorithm & Precision & Recall & F1-score & G-mean \\
        \hline
        SVM & 0.9995 & 0.7246 & 0.8402 & 0.8510 \\
        \hline
        Class-weighted SVM & 0.9877 & 0.7778 & 0.8703 & 0.8765 \\
        \hline
        RUS-SVM & 0.9278 & 0.9034 & 0.9154 & 0.9155 \\
        \hline
        SMOTE-SVM & 0.9823 & 0.8261 & 0.8975 & 0.9008 \\
        \hline
        MM-Smote & 0.9734 & 0.9034 & 0.9371 & 0.9378 \\
        \hline
    \end{tabular}
    \label{tab:ratio10}
\end{table}

\begin{table}[H]
    \centering
    \caption{Performance comparison on imbalanced datasets with 70:1 ratio }
    \begin{tabular}{|c|c|c|c|c|}
        \hline
        Algorithm & Precision & Recall & F1-score & G-mean \\
        \hline
        SVM & 0.9999 & 0.6570 & 0.7929 & 0.8105 \\
        \hline
        Class-weighted SVM & 0.9946 & 0.7101 & 0.8286 & 0.8404 \\
        \hline
        RUS-SVM & 0.9141 & 0.9034 & 0.9087 & 0.9087 \\
        \hline
        SMOTE-SVM & 0.9914 & 0.8261 & 0.9012 & 0.9050 \\
        \hline
        MM-Smote & 0.9926 & 0.8647 & 0.9243 & 0.9264 \\
        \hline
    \end{tabular}
    \label{tab:ratio70}
\end{table}

\section{Conclusion}\label{conclusion}
This paper introduces a novel adaptive oversampling approach named MM-SMOTE. Initially, it identifies crucial minority samples(i.e, support vectors) by using slack variables of a basic soft-margin SVM classifier; after recognizing that these support vectors influence the decision hyperplane differently, our method weights each support vector based on their distances to the hyperplane; then, an adaptive oversampling is conducted by considering the k-nearest neighbors of support vectors, reflecting the local density contrast between majorities and minorities. This strategy effectively generates synthetic samples in low-density regions of majorities, increasing minority diversity and modifying the decision hyperplane. Finally, the method employs an augmented kernel function to derive a new decision function to make classification. Experimental results demonstrate the proposed method's effectiveness and stability across the financial fraud dataset with distinct imbalance ratios.

Although our method makes great use of the advantages of SVM, it also retains the disadvantage that it cannot adapt to large-scale data calculation, making it impossible for us to test on large-scale data sets. In the future, to improve this method, our research direction will focus on calculation acceleration and multi-classification problems.

\end{document}